\documentclass[letterpaper, 10 pt, conference]{ieeeconf}  

\IEEEoverridecommandlockouts                              

\usepackage[utf8]{inputenc} 
\usepackage[T1]{fontenc}    
\usepackage{url}            
\usepackage{booktabs}       
\usepackage{amsfonts}       
\usepackage{nicefrac}       
\usepackage{microtype}      
\usepackage{xcolor}         
\usepackage{graphicx} 
\usepackage{amssymb}
\usepackage{amsmath}
\usepackage{enumerate}
\usepackage{wrapfig}
\usepackage{diagbox}
\usepackage{tcolorbox}
\usepackage{caption}
\usepackage[colorlinks=true]{hyperref}
\hypersetup{
    linkcolor=red,
    citecolor=blue,
    urlcolor=blue
}

\newcommand{\transition}{P_\mathcal{T}}

\title{\LARGE \bf
SignBot: Learning Human-to-Humanoid Sign Language Interaction
}

\author{Guanren Qiao$^{1}$, Sixu Lin$^{1}$, Ronglai Zuo$^{2}$, Zhizheng Wu$^{1}$, Kui Jia$^{1}$, Guiliang Liu$^{1,*}$
\thanks{$*$ Corresponding Author}%
\thanks{${1}$ School of Data Science, the Chinese University of Hong Kong, Shenzhen (E-mail: guanrenqiao1@link.cuhk.edu.cn; liuguiliang@cuhk.edu.cn)}%
\thanks{${2}$ Imperial College London}%
}

\begin{document}

\maketitle
\thispagestyle{empty}
\pagestyle{empty}

\begin{abstract}

Sign language is a natural and visual form of language that uses movements and expressions to convey meaning, serving as a crucial means of communication for individuals who are deaf or hard-of-hearing (DHH).
However, the number of people proficient in sign language remains limited, highlighting the need for technological advancements to bridge communication gaps and foster interactions with minorities.
Based on recent advancements in embodied humanoid robots, we propose SignBot, a novel framework for human-robot sign language interaction. SignBot integrates a cerebellum-inspired motion control component and a cerebral-oriented module for comprehension and interaction. Specifically, SignBot consists of: 1) Motion Retargeting, which converts human sign language datasets into robot-compatible kinematics; 2) Motion Control, which leverages a learning-based paradigm to develop a robust humanoid control policy for tracking sign language gestures; and 3) Generative Interaction, which incorporates translator, responder, and generator of sign language, thereby enabling natural and effective communication between robots and humans. Simulation and real-world experimental results demonstrate that SignBot can effectively facilitate human-robot interaction and perform sign language motions with diverse robots and datasets. SignBot represents a significant advancement in automatic sign language interaction on embodied humanoid robot platforms, providing a promising solution to improve communication accessibility for the DHH community. Please refer to our webpage: \url{https://qiaoguanren.github.io/SignBot-demo/}

\end{abstract}

\section{INTRODUCTION}


Sign language, as the primary linguistic medium for the deaf and hard-of-hearing (DHH) communities, plays a vital role in bridging the communication barriers between these communities and others. Recent advancements in computer vision and large language models (LLMs) have significantly enhanced sign language applications, including generation, translation, and recognition~\cite{jiao2024visual, yu2024signavatars}. These advancements enable effective translation between sign language text, videos, and mesh representations. However, despite these advancements, their real-world impact on assisting individuals with disabilities remains limited. A key reason is that these systems are primarily demonstrated in models and cannot facilitate physical interaction with people in real-world scenarios.


To address this gap, Embodied Artificial Intelligence (EAI) integrates AI models into physical agents, enabling real-world interaction, task execution, and continuous learning. Recent progress in humanoid robots~\cite{gu2025humanoidsurvey} highlights their potential, as human-like structures allow seamless integration into daily environments for tasks such as housekeeping, cooking, and navigation, thereby fostering natural human–robot interaction.
\begin{figure}[htbp]
    \centering
    \includegraphics[width=1.0\linewidth]{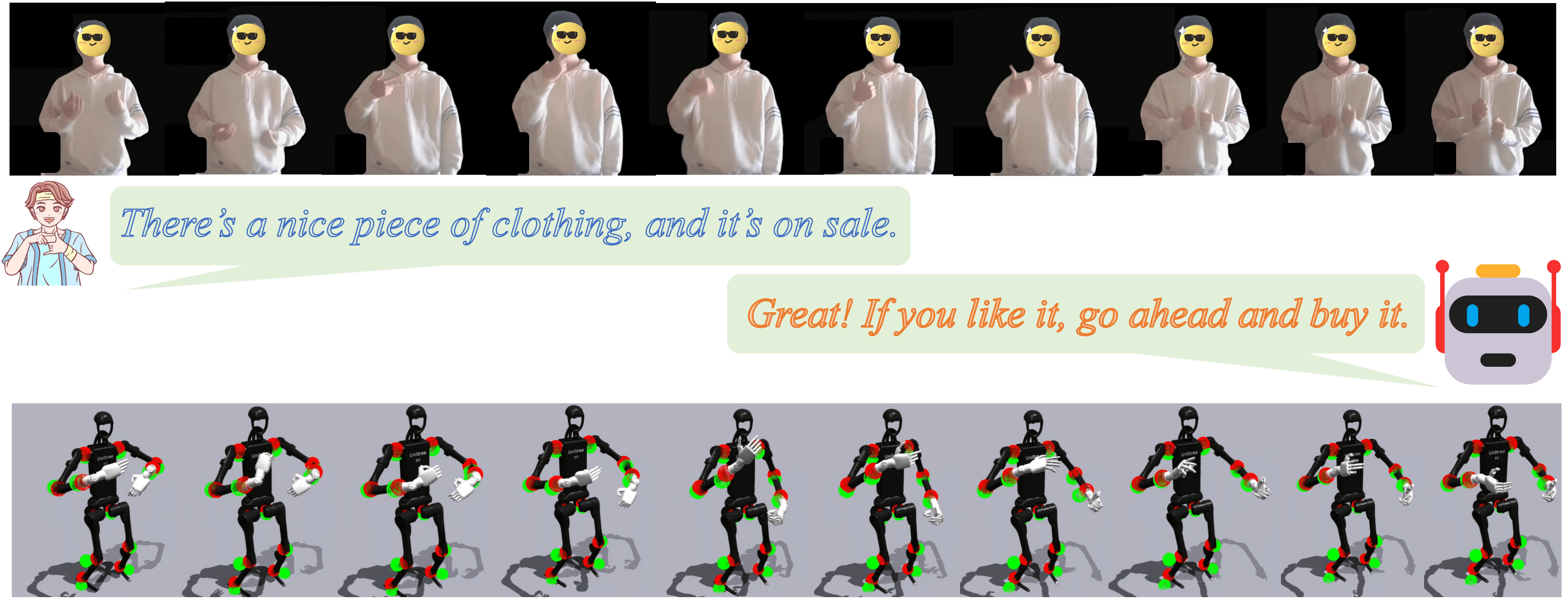}
    \caption{Motivation of SignBot: Human-Robot Sign Language Interaction.}
    \vspace{-0.3in}
    \label{fig:intro}
\end{figure}
However, no previous study has explored some methods for sign language applications.
Teleoperation-based approaches \cite{fu2024humanplus, he2024omnih2o} typically rely on human manipulation, preventing robots from autonomously performing sign language. Learning-based control methods \cite{cheng2024expressive, ji2024exbody2} primarily focus on the robot’s body without addressing the complexities of dexterous hand movements.
Additionally, many dexterous robotic hands have limited degrees of freedom (DoFs), and the lack of wrist flexibility further restricts the accurate expression of the rich and diverse movements required for sign language.

To overcome these challenges, we introduce SignBot, an expressive robotic sign language framework designed for seamless interaction with sign language users. Figure \ref{fig:intro} illustrates the motivation of our work.
SignBot mainly consists of three key components: 1) \textit{Motion Retargeting}, which maps the action sequences from human sign language datasets into a format compatible with robotic kinematics \cite{cheng2024expressive, qin2023anyteleop}. 2) \textit{Policy Training (SignBot's Cerebellum)}, which enables humanoid robots to first learn diverse sign language motions in a simulated environment with a decoupled policy \cite{he2024omnih2o, cheng2024expressive}. Specifically, we utilize decoupled body policies to learn the entire sign language gesture. The upper body, including the hands, learns to track the target sign language poses through imitation learning, while the lower body maintains a stable standing posture using an RL policy. 
3) \textit{Sign Language Interaction (SignBot's Cerebral)} integrates a sign language translator \cite{li2025uni}, a sign language responder \cite{deepseek2025deepseek}, and a sign language generator, enabling the robot to understand user expressions and respond appropriately in sign language. By combining these elements, our framework enhances real-time human-robot interaction, bridging the communication gap between sign language users and embodied robot systems.

We design various experiments to verify the performance of SignBot. Experimental results show that SignBot exhibits {\it accuracy}, {\it generalization}, {\it naturalness}, and {\it interactivity} with various datasets and robots. Overall, the main contributions of our paper are as follows:
\begin{enumerate}[(1)]
    \item \textbf{Human-Robot Interaction for Minority.} We propose an interactive sign language framework that enables seamless communication between robots and the DHH community.
    \item \textbf{Precise Sign Language Execution.} Our robot control policy robustly adapts to a diverse range of human sign language motions, ensuring stable and accurate execution.
    \item \textbf{Embodiment and Domain Adaptation.} Signbot can be transferred to different robots, achieving sign language interaction in Sim-to-Real scenarios.
\end{enumerate}

\section{RELATED WORK}

{\bf Humanoid Robotic Imitation of Human Behavior.}
For the imitating human behavior problem of humanoid robots, researchers often adopt a whole-body control learning paradigm \cite{gu2025humanoidsurvey}. This paradigm consists mainly of two approaches. One approach is to decouple the upper and lower body policies, which are responsible for managing different parts of a humanoid robot. Representative works include Exbody \cite{cheng2024expressive}, OmniH2O \cite{he2024omnih2o}, etc. Although upper and lower body policies are decoupled, they can still be integrated into a whole-body control paradigm. The alternative approach involves providing reference motions for the humanoid robot. Given the physical similarities to humans, a promising reference is the collection of human movements from motion datasets, such as Exbody \cite{cheng2024expressive}, ASAP \cite{he2025asap}, H2O\cite{he2024learning}, etc. We simultaneously leverage the advantages of both methods to perform sign language. These reference motions provide rich signals for humanoid robots to imitate human-like motions. 

{\bf Sign Language Processing.}
The field encompasses two primary research directions: sign language translation (SLT), and sign language generation (SLG). SLT and SLG form complementary pathways for bidirectional communication between deaf and hearing populations, specializing in sign-to-text and text-to-sign conversion, respectively. Some studies successfully incorporated language models (LMs) pre-trained on extensive natural language corpora into SLT frameworks, yielding substantial accuracy enhancements \cite{jiao2024visual}. Recent state-of-the art SLG works can be categorized into two classes: the first group of methods \cite{baltatzis2024neural} employs diffusion models to generate sign motions conditioned on text inputs; the second group of methods \cite{yu2024signavatars} considers the linguistic nature
of sign languages and adopts a tokenizer-LM autoregressive generation approach. 

\section{Problem Definition}\label{sec:preliminaries}

\noindent{\bf Robot Learning Environment.}
We formulate the task of tracking human sign language motions as a Partially Observable Markov Decision Process (POMDP) defined by the tuple $\mathcal{M} = (\mathcal{O}, \mathcal{S}, \mathcal{A}, \transition, \mathcal{R}, \mu_0, \gamma)$, where: 1) Within the observation space $\mathcal{O}$, each observation $o_t \in \mathcal{O}$ consists of two components: proprioception ($o_t^p$) and goal imitation ($o_t^y$). The proprioception $o_t^p$ includes essential motion-related information such as the root state, joint positions, and joint velocities. Meanwhile, the goal imitation $o_t^y$ represents a unified encoding of the whole-body sign language pose that must be tracked during RL training.
2) $s_t \in \mathcal{S}$ records the complete information and environment of the robot. We summarize a state as $s_t = [o_t,o_{t-1},...,o_{t-H}]$ and each $o_t = [o_t^p,o_t^y]$. 
2) $a_t \in \mathcal{A}$ denotes the action space, and action $a \in \mathcal{A}$ denotes the target joint positions that a PD controller uses to actuate the DoF. 
3) $r_t = \mathcal{R}(s_t,a_t)$ denotes the reward functions, which typically consist of penalty, regularization, and task rewards. These reward signals determine the level of optimality in the control policy, for which we provide a detailed introduction in Appendix A.1.
4) $\transition\in \Delta^{\mathcal{S}}_{\mathcal{S}\times\mathcal{A}}$ denotes the transition function as a mapping from state-action pairs to a distribution of future states. 
5) $\mu_0\in\Delta^{\mathcal{S}}$ denotes the initial state distribution.
6) $\gamma\in(0,1]$ denotes the discounting factor. 
Under this POMDP, our goal is learning a control policy
$\pi: \mathcal{S} \to \mathcal{A}$ that can maximize the discounted cumulative rewards $ \sum_{t=0}^{T-1} \gamma^t r_t$.


\noindent{\bf Human-Robot Sign Language Interaction.} The problem of sign Language Interaction can be modelled as a closed-loop system. Firstly, the robot should observe a sequence of the user's sign motions $ \boldsymbol{o}^{y_\text{human}}=[o^{y_\text{human}}_1,\dots,o^{y_\text{human}}_K]$. Then, the robot translates sign language $\boldsymbol{o}^{y_\text{human}}$ into a sequence of text description $\boldsymbol{x}\sim\mathcal{X}
$ ($\mathcal{X}$ denotes the space of text sequence) with the translation function $f_{\mathcal{T}}: \mathcal{{O}}^K\to\mathcal{{X}}$. To response, the system  must understand the intention of $\boldsymbol{x}_t$ and answer $\boldsymbol{x}^\prime_t$ with the responding function $f_{\mathcal{R}}: \mathcal{X}\to\mathcal{X}$    
. 
Based on the text response $\boldsymbol{x}_t^\prime$, the system should generate sign language $\boldsymbol{o}_t^{y_{robot}}$ with the generation function $f_{\mathcal{G}}: \mathcal{X}\to\mathcal{O}^K$, which can be used as imitation goals for robot controller. Specifically, we formulate the sign language generation problem as a conditional sequence generation task, where the goal is to generate a sign language SMPL-X sequence from input semantic information. The output sign language sequence
$\boldsymbol{o}^y$ is the SMPL-X representation space of sign language and $K$ denotes the length of the motion sequence. Typically, this is modeled by the conditional probability distribution$
P_\mathcal{G}(\boldsymbol{o}^y|\boldsymbol{x}) = \prod_{k=1}^{K} P_\mathcal{G}(o^y_k \mid o^y_{<k}, \boldsymbol{x})$.

\begin{figure*}[htbp]
    \centering
    \includegraphics[width=1.0\linewidth]{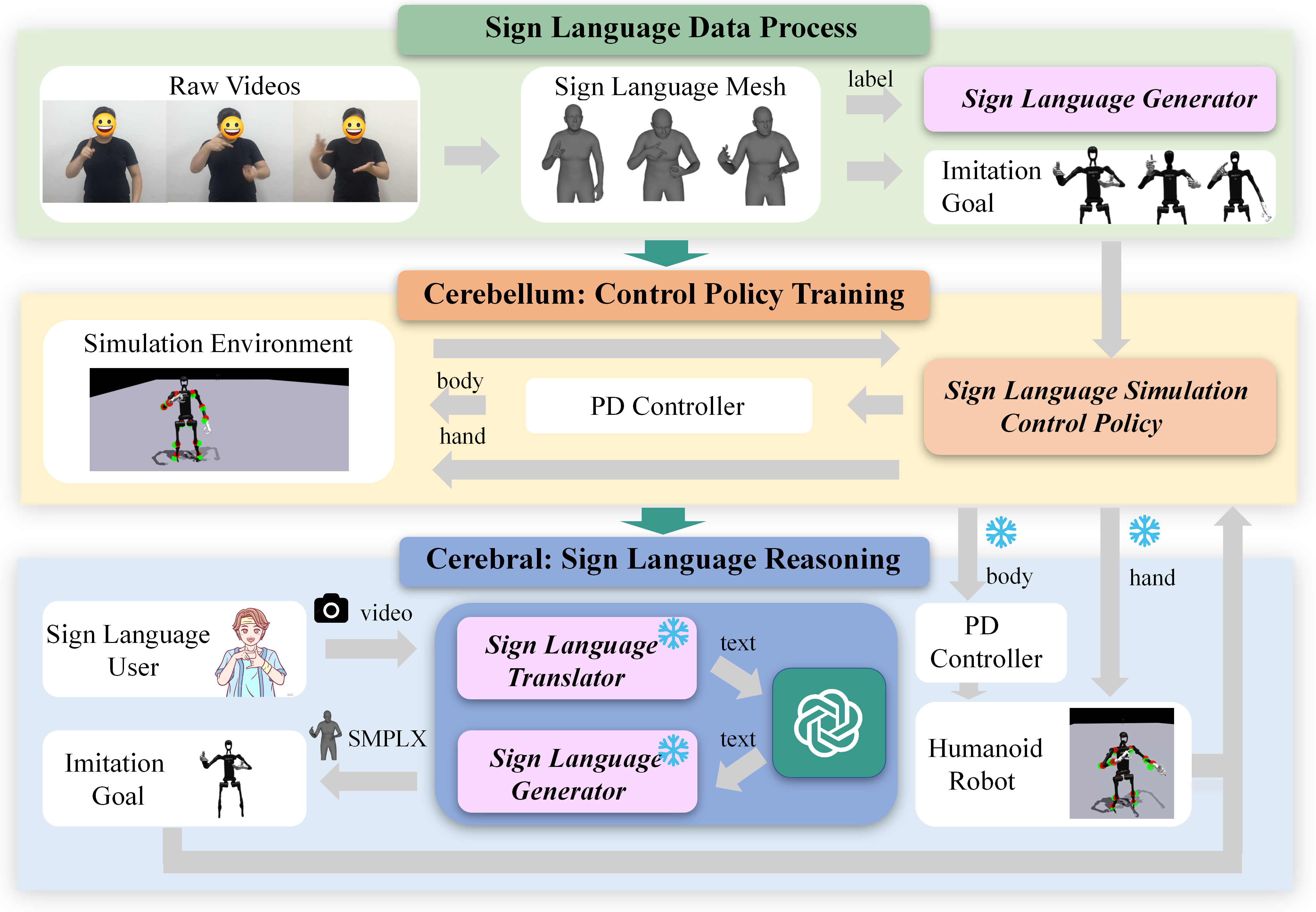}
    \caption{ Overview of SignBot: The framework consists of three stages: (1) \textit{Motion Retargeting} aligns human sign language gestures with the body structure of humanoid robots (Section~\ref{sec:motion_retargeting}). In addition, we use the processed mesh along with text labels to train the sign language generator model. (2) {\it Cerebellum} performs Sim2Real policy training that enables the robot to track various sign language gestures in the simulated environment and deploy the policy to real-world (Section~\ref{sec:cerebellar}). (3) {\it Cerebral} conducts sign language reasoning to facilitate communication with sign language users through the sign language translator, response, and generator within the cerebral (Section~\ref{sec:brain}).}
    \vspace{-0.3in}
    \label{fig:pipeline}
\end{figure*}

\section{SignBot}\label{sec: signbot}

In this section, we introduce the pipeline of SignBot, which is divided into three parts: 1) \textbf{Motion Retargeting} of the body and hands, 2) \textbf{Policy Training} to control the robot's movements as \textit{"SignBot's cerebellum"}, and 3) \textbf{Sign Language Reasoning} for comprehensive and responding users' sign languages as \textit{"SignBot's cerebral"}. Figure \ref{fig:pipeline} illustrates the SignBot pipeline.
\subsection{Motion Retargeting}\label{sec:motion_retargeting}

As shown in the first stage of Figure \ref{fig:pipeline}, we extract the motion from the video mesh for subsequent data processing. To mitigate differences in body shape between humans and the humanoid robot, we perform retargeting separately for the human body and hands.

\noindent{\bf Body Retargeting.} Our humanoid body retargeting system is based on the \cite{peng2022ase}. By establishing a mapping relationship between the source keypoints and the target keypoints, we follow a dual T-Pose (the standard poses of the source and target skeletons) as a spatial alignment reference. We convert the local quaternion of each joint to the form of axis angles. It is important to note that to make the robot's sign language movements more natural, we add two additional degrees of freedom at the robot's wrist and transform the wrists and shoulders represented by a 1D joint to a 3D joint. 



\noindent{\bf Hand Retargeting.} To map human hand pose data to the joint positions of the Linker hand (the robot hands used by SignBot), we apply the preprocessing method from \cite{qin2023anyteleop}, adapting it specifically for the Linker hand.  This process is often formulated as an optimization problem, where the difference between the keypoint vectors of the human hand model and the dexterous hand is minimized. Since the linker hand is larger than a typical dexterous hand, we adjust the scale factor in the optimization process. Additionally, we modify the regularization term to smooth the sign language movements between consecutive frames.

 \subsection{SignBot's Cerebellar: Control Policy Training} \label{sec:cerebellar}

Within our SignBot humanoid agent, the cerebellum controls the low-level movements for performing sign language.
As shown in the second stage of Figure \ref{fig:pipeline}, we train a control policy to enable the humanoid robot to track and imitate sign language gestures in a simulated environment. In this section, we discuss our approach from three key perspectives: observation space, decoupled policy, and reward design.

\noindent{\bf Observation Space.} SignBot's observation $o_t \in \mathcal{O}$ consists of proprioception ($o_t^p$) and goal imitation ($o_t^y$) (Section~\ref{sec:preliminaries}). Our proprioception is defined as $o_t^p \triangleq \left[\mathbf{q}_t, \dot{\mathbf{q}_t}, \mathbf{v}_t, \mathbf{w}_t, \mathbf{g}_t, \mathbf{a}_{t-1}\right]$, which includes joint position $\mathbf{q}_t \in \mathbb{R}^{55}$ (DoF position), joint velocity $\dot{\mathbf{q}}_t \in \mathbb{R}^{55}$ (DoF velocity), root linear velocity $\mathbf{v}_t \in \mathbb{R}^{3}$, root angular velocity $\mathbf{w}_t \in \mathbb{R}^{3}$, root projected gravity $\mathbf{g}_t \in \mathbb{R}^{3}$, and the previous action $\mathbf{a}_{t-1} \in \mathbb{R}^{55}$. The goal observation is $\mathbf{o_t^y} \triangleq \left[\hat{\mathbf{q}}^{kp},\hat{\mathbf{q}}, \dot{\hat{\mathbf{q}}}\right]$, where $\hat{\mathbf{q}}^{kp} \in \mathbb{R}^{14 \times 3}$ are the positions of 14 selected reference keypoints \cite{cheng2024expressive} to ensure that the humanoid robot and the imitation goal are oriented in the same direction, $\hat{\mathbf{q}} \in \mathbb{R}^{55 \times 3}$ are the positions of all reference joints, and $\dot{\hat{\mathbf{q}}}$ is the linear velocity of the reference joints. At the same time, we generally perform domain randomization \cite{he2024omnih2o} in the simulation environment to ensure robustness.


\noindent{\bf Decoupled Policy.} Given that sign language involves precise coordination of both hand poses and full-body motion with high DoFs, learning a unified control policy is inherently challenging. More importantly, while the dual arms in the upper body can often be governed by a shared control strategy, the control approach for the lower body may vary significantly depending on the physical design of the humanoid robot. For instance, bipedal humanoid robots typically utilize RL controller, whereas wheeled robots often employ model predictive control (MPC) techniques.

Motivated by recent advances in whole-body humanoid control \cite{cheng2024expressive}, we adopt a decoupled architecture that separates the control policies of the upper and lower body.The primary objective of the upper-body policy \( \pi^{upper} \) is to track the retargeted actions, while the lower-body policy \( \pi^{lower} \) ensures balance in the robot’s default standing pose while adapting to the movements of \( \pi^{upper} \). 
\begin{align}
    &\pi^{upper} = {\arg\min}_{\pi}\mathbb{E}_{(s,a^{up}_*)}\Big[\mathcal{D}_f[\delta(a_*^{up})\|\pi(a^{up}|s)]\Big]\\
    & \pi^{lower} = {\arg\max}_{\pi} \mathbb{E}_{\pi(a^{low}|s)}\left[\sum_{t=0}^\infty \gamma^t r_t(s,a)|a^{lower}\sim\pi^{upper}\right]\nonumber
\end{align}
Where:  
1) \( a_*^{up} \) represents the retargeted action of the upper body, and \( \delta \) denotes the Dirac delta function.  
2) \( \mathcal{D}_f(\cdot\|\cdot) \) refers to the \( f \)-divergence (e.g., KL-divergence) between two distributions.  
3) The whole-body humanoid action, \( a = [a^{low}, a^{up}] \), is the concatenation of upper and lower body actions. 4) The reward from the humanoid control learning environment can be represented by the weighted penalty $r_\mathfrak{P}$, task $r_\mathfrak{T}$, and regularization $r_\mathfrak{R}$ terms: 
$r_t = \beta_\mathfrak{T}r_\mathfrak{T}+\beta_\mathfrak{P}r_\mathfrak{P}+\beta_\mathfrak{R}r_\mathfrak{R}$. In our task, task rewards ($r_\mathfrak{T}$) measure the robot's performance in tracking joint motions and body velocities. Penalty rewards ($r_\mathfrak{P}$) serve to discourage undesirable outcomes such as falling and violating dynamic constraints like joint or torque limits.
Regularization rewards ($r_\mathfrak{R}$) are used to align the humanoid's sign language gestures with human preferences. 
Appendix A.1 specifies the reward functions.

Note that to train the lower-body policy $\pi^{lower} $, we apply the PPO \cite{schulman2017proximal} algorithm for legged humanoid robots and MPC for wheeled humanoid robots. Appendix A.2 reports the detailed parameter settings.
In this way, SignBot can be scaled to multiple embodiments, for which we demonstrate in the experiment.

\vspace{-0.1in}
\subsection{SignBot's Cerebral: Sign Language Reasoning} \label{sec:brain}
Within our SignBot humanoid agent, the cerebral system controls high-level reasoning skills, enabling it to respond to sign language and generate appropriate responses.
An ideal embodied humanoid robot needs to have both cerebellar and cerebral capabilities, enabling it to interact effectively with its surrounding environment and other agents. The third stage of Figure \ref{fig:pipeline} illustrates the cerebral system of the SignBot. This system processes the user's input and generates an appropriate sign language response, which serves as the imitation goal for the control policy ($\mathbf{o_t^c}$ in Section~\ref{sec:cerebellar}). 


\vspace{0.05in}\noindent{\bf Implementation.} 
SignBot utilizes a camera to observe the motions of sign language users and then stores this as a video input into its cerebral system. 
For communicating with sign language users in real-time, SignBot's cerebral system is implemented by three models: a \textit{sign language translator} understanding sign language contents, a \textit{sign language responder} interpreting semantics and generating responses, and a \textit{sign language generator} converting texts to the SMPLX format with Transformer-based models. We introduce these two models in the following:

\textit{Sign Language Translator} is implemented with a LLM \cite{li2025uni}. In the pre-training stage, it extracts the sign language features from sign language videos and images, aligns them with the dimensions of the language model, and then inputs them into the model. In the fine-tuning stage, we utilize the translation text from the sign language dataset~\cite{zhou2021improving, duarte2021how2sign} to construct the supervised labels for fine-tuning our translation model.

\textit{Sign Language Responder} is implemented using the DeepSeek-Chat API \cite{deepseek2025deepseek} due to its ability to seamlessly comprehend semantic information and facilitate multi-turn conversations. Other chat models can also be used here. To simulate natural conversations with sign language users, we design a suitable prompt template. When generating a response, the system integrates the text-format sign language into the prompt and produces a contextually appropriate reply. We ensure that DeepSeek is aware of the vocabulary and scope covered by the CSL dataset to prevent the output of dangerous or out-of-distribution (OOD) corpus. 


\textit{Sign Language Generator} autoregressively generates sign motions from text input based on a multilingual LM \cite{liu2020multilingual}. First, we design a decoupled VQVAE tokenizer to
map continuous sign motions to discrete tokens over upper body (UB), left hand (LH), and right hand (RH) movements. Given a $K$-frame sign sequence, we decompose it into three part-wise motion sequences based on the SMPL-X: \( \mathbf{C}^u \in \mathbb{R}^K \), where \( u \in \{\text{UB}, \text{LH}, \text{RH}\} \). For each body part, we train a separate VQ-VAE comprising an encoder that projects the sequence into a latent space $
\mathbf{C}^u_{f_{enc}} = \{ c^u_{f_{enc},k} \}_{k=1}^K \in \mathbb{R}^{K_{f_{enc}} \times C}
$, a decoder for reconstruction, and a learnable codebook \( \mathbf{Z}_u \in \mathbb{R}^{N^u_Z \times d} \), where \( N^u_Z \) represents the number of codes and \( d \) denotes the code dimension. Then, for each pose, we can derive a set of discrete tokens $\hat{\boldsymbol{z}}_{1,\dots,K}^u=[\hat{z}^u_1,\dots,\hat{z}^u_K]$, which searches for the nearest neighbor from the codebook \( Z_u \):
\begin{equation}
\begin{aligned}
    \hat{z}^u_k = \arg\min_{z_n \in \mathbf{Z}_u} \| s^u_{f_{enc},k} - z_n \|^2, \quad n \in [1, N^u_Z]
\end{aligned}
\end{equation}
where $\forall{u \in \{\text{UB}, \text{LH}, \text{RH}\} }$. Given a text description $\boldsymbol{x}$, the generator retrieves word-level signs based on the $\boldsymbol{x}$ from external dictionaries made by the decoupled tokenizer. These word-level signs are represented with discrete tokens $\hat{\boldsymbol{z}}_{1,\dots,K}^u$. We feed these tokens and text sequence $\boldsymbol{x}$ into the LM encoder at the same time. During decoding, we adopt a multi-head decoding strategy. We design three language modeling heads, implemented as fully connected layers, to predict motion tokens for each body part from $m$ simultaneously at each step.  The decoding process can  be formulated as:
 \begin{equation}
 \begin{aligned}
P_{Dec}&(\boldsymbol{o}^y|\mathbf{h}) = \prod_{k=1}^{K} \prod_{u}P_{Dec}(o_k^{y^{u}}|\boldsymbol{o}_{<k}^{y,\mathbf{h}}) \
\end{aligned}
\end{equation}
where $(\boldsymbol{o}_{<k}^{y^{u}}, \mathbf{h})$ is the simplification of $\boldsymbol{o}_{<k}^{y,\mathbf{h}}$, $\boldsymbol{o}^y = \{o_1^{y^\text{UB}}, o_1^{y^\text{LH}}, o_1^{y^\text{RH}}, \dots, o_K^{y^\text{UB}}, o_K^{y^\text{LH}}, o_K^{y^\text{RH}}\}$ and $\mathbf{h}=f_\text{Enc}(\boldsymbol{x}, \boldsymbol{\hat{z}}^u_{1,\dots,K})$ denotes the output of LM Encoder. Finally, the derived motion tokens are used to reconstruct sign motions. 


\textbf{Sim-to-Real Deployment.}
In real-world environments, robot movement from positions from 
$t$ to $t+1$ is not instantaneous. To ensure smooth transitions, we employ the Ruckig algorithm \cite{berscheid2021jerk} for online trajectory generation with third-order (jerk) constraints and complete kinematic targets. Ruckig computes time-optimal trajectories between arbitrary states, defined by position, velocity, and acceleration, while respecting velocity, acceleration, and jerk limits. To maintain smooth motion, we interpolate intermediate values between target positions. For safety during deployment, we avoid commanding joint angles near their physical limits, as slight discrepancies between simulation and reality may lead to motor power loss or low-voltage issues, even when using identical limit settings.

\section{Environment}
\label{sec: environment}
\vspace{-0.05in}
\noindent{\bf Experiment Settings.} To conduct a comprehensive evaluation, we quantify the performance of SignBot in simulated (IssacGym~\cite{makoviychuk2021isaac}) and realistic environments from the following perspectives:
1) {\it Accuracy}: How to effectively align sign language actions between robots and humans?
2) {\it Generalization}: How well does SignBot perform with diverse sign language datasets and different robots?
3) {\it Naturalness}: How effectively does SignBot imitate human-like sign language norms in the real-world interaction scenarios?
4) {\it Interactivity}: What is the DHH community’s genuine evaluation of this work?
We use the public CSL-Daily \cite{zhou2021improving} and How2Sign (ASL)~\cite{duarte2021how2sign} Sign Language dataset. Experiments span different embodiments: H1 legged robot, W1 wheeled robot, and Linker hand.

\noindent{\bf Metrics.} We adopt consistent metrics across settings, including errors in 1) DoF positions, body yaw, linear velocity, and roll/pitch. 2) \textit{Cumulative Rewards}: are calculated by all the weighted reward functions. 3) \textit{BLEU-1} calculates the single-word overlap between the generated text and the reference. 4) \textit{BLEU-4} evaluates the match of four-word sequences (4-grams). 5) \textit{ROUGE} assesses how well the generated output covers the essential content of the reference. 6) \textit{DTW-PA-JPE} \cite{baltatzis2024neural} evaluates sequence-level distances between the generated signs and ground truth.



\noindent{\bf Comparison Methods.} To demonstrate the effectiveness of SignBot, we compare with other baselines based on whole-body control or RL as follows: 1) \textbf{SignBot (w/o Lower-Body Tracking} follows SignBot, but allows the lower body to self-adapt and maintain balance. 2) \textbf{Whole-Body Tracking + AMP} uses an AMP reward \cite{peng2022ase} to encourage the transitions of the policy to be similar to the motions of the sign language features. 3) \textbf{Whole-Body Tracking} \cite{he2024omnih2o} learns the movement of the upper body and lower body simultaneously.

\subsection{Accuracy: Alignment of Sign Language between Human and Robot}


Precise execution of sign language is vital for effective communication within the DHH 
community, as even minor inaccuracies can lead to misunderstandings, in contrast to boxing techniques, where approximate gesture replication suffices for demonstration purposes. To construct a high-fidelity dataset with precise 3D SMPLX annotations, we leverage state-of-the-art methods for 3D hand \cite{potamias2025wilor} and body reconstruction \cite{lin2023onestage}. Specifically, for each 2D sign language video, we first utilize OSX \cite{lin2023onestage} to obtain an initial body pose estimation. Recognizing that OSX often struggles with accurate arm and hand pose estimation, we implement a two-stage refinement process. For high-precision hand pose reconstruction, we employ WiLoR \cite{potamias2025wilor}, a cutting-edge 3D reconstruction pipeline that can robustly detect and reconstruct even challenging hand configurations. The hand pose parameters and global orientation outputs from WiLoR are then directly integrated to replace the corresponding OSX-derived estimates, ensuring high fidelity. 


As the datasets contain multiple demonstrators performing the same actions, we select the sign language motions of one demonstrator and segment them by difficulty based on the length of the label (longer labels correspond to motions of longer duration, for example, labels with 10 words, 10-20 words, and 20 words). We divide the CSL-Daily sign language dataset~\cite{zhou2021improving} into three difficulty levels: simple (929 sentences), intermediate (4558 sentences), and difficult (1089 sentences). We compare SignBot with the previously mentioned baselines on the data across these three difficulty levels. Appendix A.3 records the running details of this experiment. 

\begin{table*}[htbp]
\centering
\resizebox{0.99\textwidth}{!}{
\begin{tabular}{c|ccccc}
\hline
\diagbox{Baseline}{Metric} & DoF Pos $\downarrow$ & Yaw $\downarrow$ & Linear Velocity $\downarrow$ & Roll\&Pitch $\downarrow$ & Cumulative Rewards $\uparrow$ \\

\hline
\multicolumn{6}{c}{Easy} \\
\hline
Whole-Body Tracking & $0.97\pm0.01$&$ 0.16 \pm 0.01$& $\boldsymbol{0.18  \pm 0.01 }$&$ 0.05 \pm 0.00$& $ 10.41 \pm 0.70$ \\
Whole-Body Tracking + AMP & $1.13\pm 0.01$ & $0.11\pm 0.00$ & $0.22\pm 0.00$ & $0.04\pm 0.01$ & $8.89\pm 0.32$\\
SignBot (w/o Lower-Body Tracking) &$0.85\pm0.01$&$0.95\pm0.01$&$1.86\pm0.02$&$0.26\pm0.01$&$0.54\pm0.01$\\
SignBot & $\boldsymbol{0.59 \pm 0.01}$&$ \boldsymbol{0.04 \pm 0.00}$&$ 0.20 \pm 0.01$&$ \boldsymbol{0.03 \pm 0.01}$&$ \boldsymbol{12.46 \pm 0.40}$\\
\hline
\multicolumn{6}{c}{Medium} \\
\hline
Whole-Body Tracking & $0.95\pm0.01$&$ 0.17 \pm 0.00$& $\boldsymbol{0.17  \pm 0.01 }$&$ 0.05 \pm 0.00$& $ 10.63 \pm 0.63$ \\
Whole-Body Tracking + AMP & $1.12\pm 0.01$ & $0.10\pm 0.00$ & $0.22\pm 0.00$ & $\boldsymbol{0.04\pm 0.00}$ & $9.18\pm 0.32$\\
SignBot (w/o Lower-Body Tracking) &$0.87\pm0.01$&$0.97\pm0.00$&$1.87\pm0.02$&$0.27\pm0.01$&$0.33\pm0.01$\\
SignBot & $\boldsymbol{0.61 \pm 0.00}$&$ \boldsymbol{0.08 \pm 0.00}$&$ 0.20 \pm 0.01$&$ \boldsymbol{0.04 \pm 0.01}$&$ \boldsymbol{12.24 \pm 0.57}$\\
\hline
\multicolumn{6}{c}{Hard} \\
\hline
Whole-Body Tracking & $0.94\pm0.01$&$ 0.17 \pm 0.00$& $\boldsymbol{0.17  \pm 0.01 }$&$ 0.05 \pm 0.00$& $ 9.67 \pm 0.20$ \\
Whole-Body Tracking + AMP & $1.11\pm 0.02$ & $0.10\pm 0.00$ & $0.23\pm 0.01$ & $0.04\pm 0.01$ & $9.04\pm 0.24$\\
SignBot (w/o Lower-Body Tracking) &$0.85\pm0.00$&$0.92\pm0.01$&$2.00\pm0.02$&$0.04\pm0.00$&$0.52\pm0.03$\\
SignBot & $\boldsymbol{0.59 \pm 0.01}$&$ \boldsymbol{0.04 \pm 0.00}$&$ 0.20 \pm 0.03$&$ \boldsymbol{0.03\pm 0.01}$&$ \boldsymbol{12.56 \pm 0.30}$\\
\hline
\end{tabular}
}
\caption{Tracking Performance: We compare model performance under three difficulty levels. Bolded results indicate the best performance. The rewards calculation denotes the cumulative returns over a trajectory, while the other metrics are calculated using the mean square error.}
\label{tab:simulation performance}
\vspace{-0.2in}
\end{table*}

Notably, since the Unitree H1's wrist has only one DoF, it can only rotate while maintaining a fixed orientation and cannot bend. To address this, we modify the robot by adding two additional DoFs to the wrist, which allows the wrist to move up/down and left/right, enabling more flexible sign language gestures. Table \ref{tab:simulation performance} illustrates the training performance of each baseline under different difficulty levels. The results indicate that SignBot achieves the lowest error in metrics such as DoF position tracking and yaw angle tracking, while also achieving the highest reward value, demonstrating the effectiveness of the SignBot control policy. The improvement in tracking accuracy and the reduction in training difficulty are attributed to the provision of upper body DoF position data and the tracking of lower body bent standing position. SignBot excels in controlling yaw and roll\&pitch angles, indicating strong stability of the robot base. We also test other baselines, such as tracking the upper and whole body DoF positions, but the performance of these is unsatisfactory. This is because sign language differs from other upper-body movements; it is flexible and varied, and the frequency of sign language actions in the dataset is relatively fast. In addition, we observe that the length of the sentences does not significantly affect SignBot's performance. When the sentences are short, SignBot tends to have larger errors in tracking keypoints. This may be due to the shorter episode length, which causes the robot to complete and reset its random learning for the next action quickly. During this period, it needs to adjust the robot's global pose frequently. 



\subsection{Generalization: Imitation across diverse datasets and robots}

\begin{figure}[htbp]

    \includegraphics[width=1.0\linewidth]{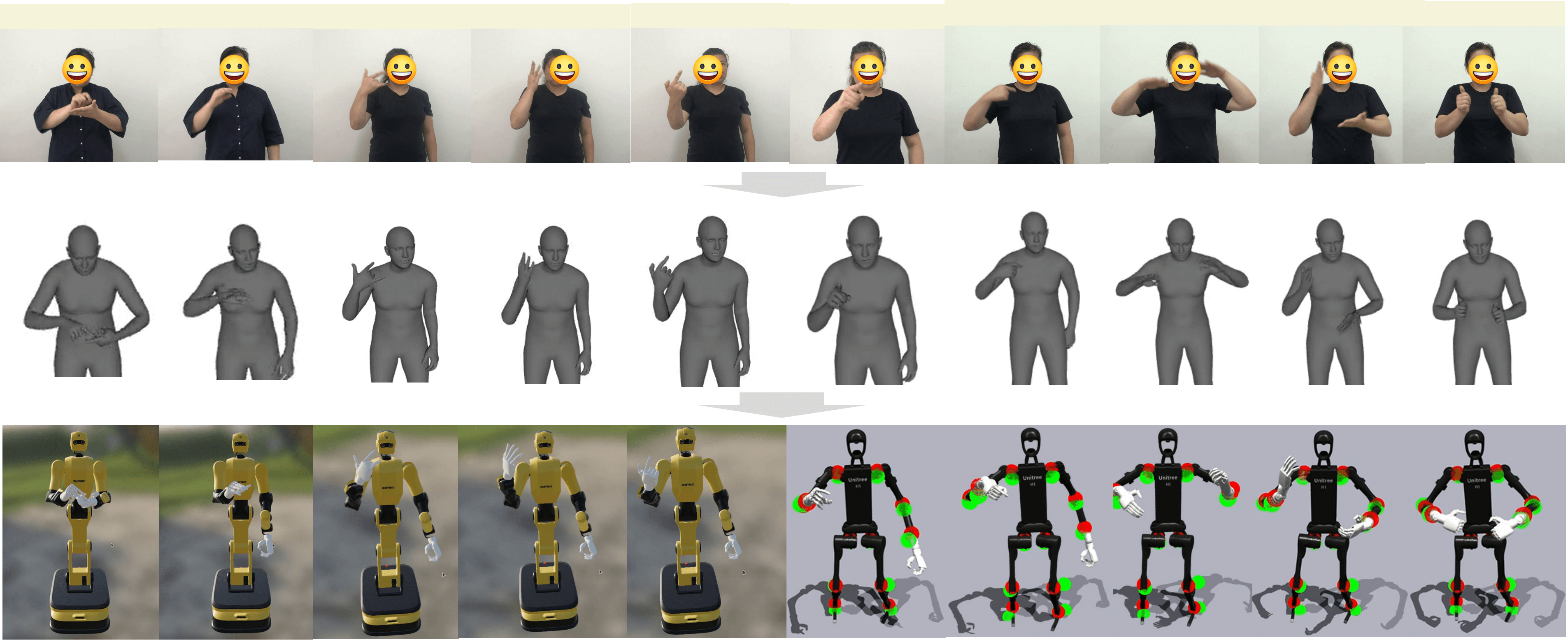}
    \caption{ Sign Language Alignment between Human and Robots: We display the source video of human sign language in the first row, followed by the mesh from video processing, and the last row shows the results of different robots. For the H1 robot, the \textcolor{red}{red} nodes represent the robot dof pos, while the \textcolor{green}{green} nodes represent the retargeted demonstration nodes.}\label{fig:w1_acc}
    \vspace{-0.1in}
\end{figure}

{\bf Generalizing SignBot to different embodiments.} We retarget the human sign language data to align with the joints of humanoid robots to visualize the precision in imitating sign language poses. Figure \ref{fig:w1_acc} illustrates the alignment between human and robot sign language gestures after applying our SignBot method, demonstrated on both the legged Unitree H1 and the wheeled W1 robots. A notable observation is that SignBot maintains a high accuracy in imitating sign language across both robots.

{\bf Generalizing SignBot to different Datasets.}To demonstrate the generalization capability of SignBot, we also evaluate SignBot with the How2Sign dataset. Specifically, we preprocess a portion of the 2,286 data entries and evaluate the robustness of the policy for different sign languages following Table \ref{tab:simulation performance}. Table \ref{tab:simulation performance2} illustrates SignBot's performance on the How2Sign dataset, demonstrating that SignBot achieves a minimal tracking error. 

\begin{table}[htbp]
\centering
\resizebox{0.49\textwidth}{!}{
\begin{tabular}{c|cccc}
\hline
\diagbox{Baseline}{Metric} & DoF Pos $\downarrow$ & Yaw $\downarrow$ & Linear Velocity $\downarrow$ & Roll\&Pitch $\downarrow$ \\
\hline
\multicolumn{5}{c}{Easy} \\
\hline
Whole-Body Tracking & $1.05\pm0.02$ & $0.32 \pm 0.00$ & $\boldsymbol{0.16 \pm 0.01}$ & $\boldsymbol{0.04 \pm 0.00}$ \\
SignBot & $\boldsymbol{0.63 \pm 0.02}$ & $\boldsymbol{0.07 \pm 0.02}$ & $0.19 \pm 0.03$ & $0.07 \pm 0.01$ \\
\hline
\multicolumn{5}{c}{Medium} \\
\hline
Whole-Body Tracking & $0.88\pm0.02$ & $0.34 \pm 0.01$ & $\boldsymbol{0.15 \pm 0.02}$ & $\boldsymbol{0.04 \pm 0.01}$ \\
SignBot & $\boldsymbol{0.57 \pm 0.02}$ & $\boldsymbol{0.11 \pm 0.01}$ & $0.17 \pm 0.02$ & $0.05 \pm 0.00$ \\
\hline
\multicolumn{5}{c}{Hard} \\
\hline
Whole-Body Tracking & $0.90\pm0.03$ & $0.30 \pm 0.03$ & $\boldsymbol{0.17 \pm 0.02}$ & $\boldsymbol{0.05 \pm 0.01}$ \\
SignBot & $\boldsymbol{0.56 \pm 0.01}$ & $\boldsymbol{0.11 \pm 0.02}$ & $0.18 \pm 0.02$ & $\boldsymbol{0.05 \pm 0.01}$ \\
\hline
\end{tabular}
}
\caption{Generalization Experiment.}
\label{tab:simulation performance2}
\vspace{-0.2in}
\end{table}

\subsection{Naturalness: Sim-to-Real Human-Robot Interaction}

\begin{figure*}[htbp]
    \centering
    \includegraphics[width=1.0\linewidth]{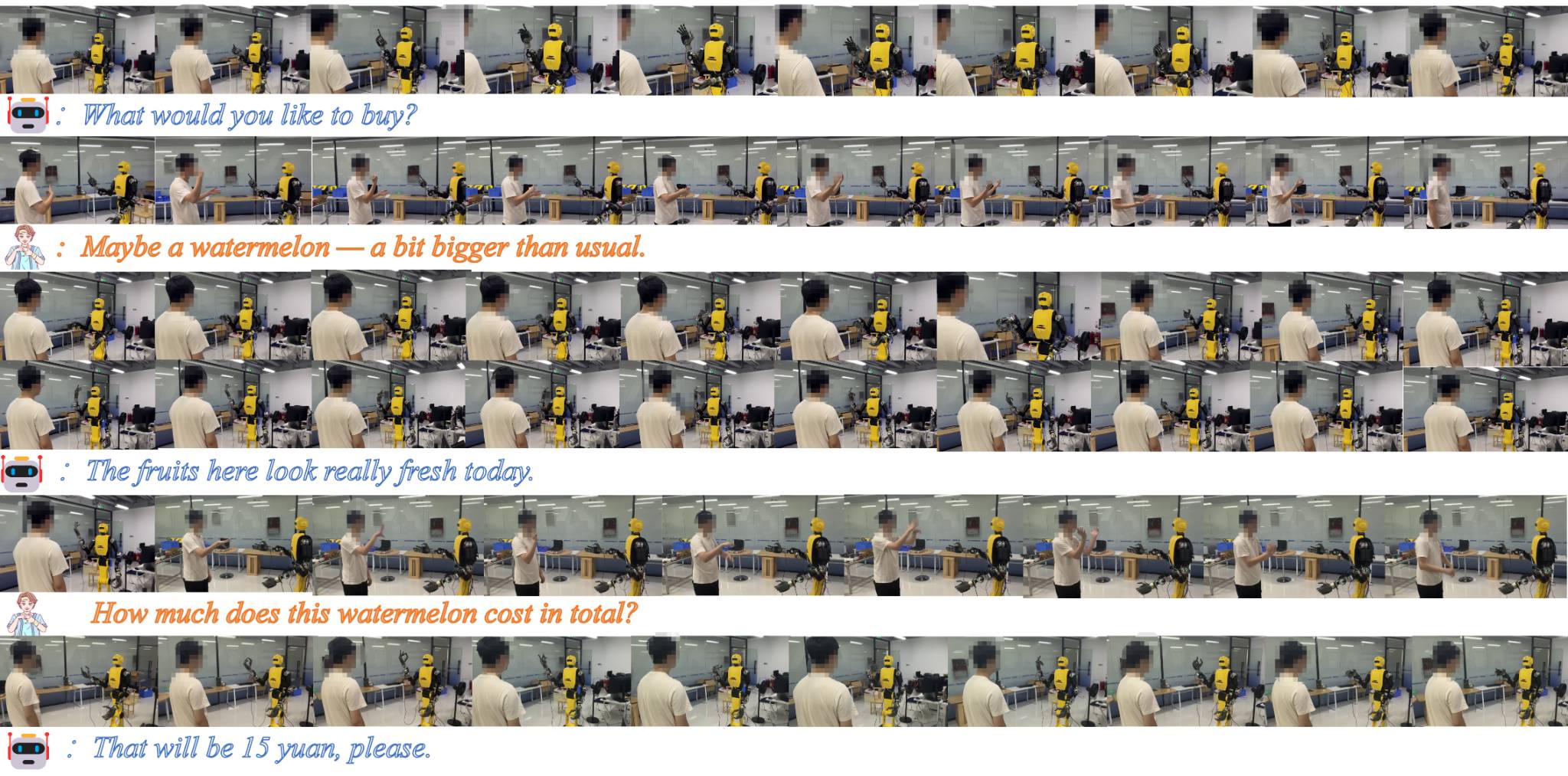}
    \vspace{-0.2in}
    \caption{ An example of real-world interaction between the robot and the human customer.}
    \label{fig:interaction}
    \vspace{-0.1in}
\end{figure*}

To demonstrate the naturalness of SignBot, we choose the W1 robot to demonstrate the performance of the SignBot framework in realistic environments. The W1 robot and Linker Hand are controlled through the ROS system to drive the various joints during real deployment. Figure \ref{fig:interaction} illustrates several examples of interactions between a human and a robot functioning as a supermarket cashier. The individual shown in the figure is a proficient sign language user from our research team. In this scenario, the robot initiates communication by asking customers about their intended purchases. More examples can be found in the supplementary video material.



\begin{table*}[htbp]
\centering
\resizebox{1.0\textwidth}{!}{
\begin{tabular}{c|ccc|c|cc}
\hline
\textbf{Translation Method} & \textbf{BLEU-1} $\uparrow$ & \textbf{BLEU-4} $\uparrow$ & \textbf{ROUGE} $\uparrow$ &\textbf{Generation Method} & \textbf{DTW-PA-JPE (body)} $\downarrow$ & \textbf{DTW-PA-JPE (hand)} $\downarrow$ \\
\hline
 C$^2$RL (CSL) \cite{chen2024c2rl} & 49.32 & 21.61 & 48.21 & NSA (ASL) \cite{baltatzis2024neural} & 7.83 & 7.33 \\
\hline
Translator (ASL) & 40.20 & 14.90 & 36.01 & Generator (ASL) & 6.82 & 2.35 \\
\hline
Translator (CSL) & 55.08 & 26.36 & 56.51 & Generator (CSL) & 6.24 & 1.71 \\
\hline
Translator (CSL, Deployment) & 53.20 & 20.53 & 54.48 & Generator (CSL, Deployment) & 7.63 & 2.19 \\
\hline
\end{tabular}
}
\caption{Evaluation of Sign Language Translation and Generation Models in the simulation and real environment.}
\label{tab:sign_language_models}
 \vspace{-0.1in}
\end{table*}

We evaluate the performance of the sign language translation and generation modules in interaction. We compare our modules with the current SOTA baselines on the ASL and CSL datasets using the test sets (since the translator is a gloss-free method, we compare it with the gloss-free baseline instead of the gloss-based baseline). Additionally, we test the effectiveness of both modules using real-world sign language videos and dialogue corpus. Since our sign language users are familiar with CSL, the actual performance is only evaluated on CSL. Table \ref{tab:sign_language_models} shows that our modules outperform the existing SOTA baseline. "Deployment" refers to the evaluation using our real test data from non-CSL test sets during the actual device deployment phase. Furthermore, since the dialogue content mainly falls within the vocabulary and scope covered by the CSL dataset, the performance degradation is not significant, providing a basic guarantee for sign language interaction.


\subsection{Interactivity: Human Feedback from the DHH Community}
We conduct a user study and invite members of the DHH community to evaluate this work. Due to constraints of the experimental venue and hardware safety concerns, we do not involve DHH individuals in actual interactions. As an alternative, we provide comprehensive documentation and video recordings of the entire sign language interaction process to facilitate their assessment.

\begin{table}
\centering

\resizebox{0.49\textwidth}{!}{
\begin{tabular}{l|cccccc}
\toprule
\textbf{Evaluator} & \textbf{Satisfaction} & \textbf{Enjoyment} & \textbf{Naturalness} & \textbf{Accuracy} & \textbf{Understanding} & \textbf{Convenience} \\
\midrule
Evaluator 1 & 6 & 5 & 6 & 5 & 6 & 8 \\
Evaluator 2 & 7 & 8 & 7 & 6 & 5 & 5 \\
Evaluator 3 & 8 & 7 & 9 & 7 & 7 & 8 \\
Evaluator 4 & 7 & 6 & 7 & 7 & 7 & 8 \\
Evaluator 5 & 6 & 6 & 6 & 5 & 6 & 8 \\
Evaluator 6 & 6 & 9 & 5 & 7 & 6 & 9 \\
Evaluator 7 & 9 & 9 & 7 & 8 & 6 & 5 \\
Evaluator 8 & 3 & 3 & 3 & 3 & 3 & 3 \\
Evaluator 9 & 1 & 1 & 1 & 1 & 5 & 6 \\
Evaluator 10 & 8 & 6 & 7 & 7 & 7 & 6 \\
\bottomrule

\end{tabular}
}
\caption{Evaluation Results from the DHH Community}
\label{tab: DHH_evaluation}
\vspace{-0.2in}
\end{table}

\textit{Experiment Setup}: Among these people, there are young and elderly individuals, all coming from different places of birth. To ensure fairness, we do not collect any personal or private information from the experiment participants. The assessment covers the following aspects: 1) Overall satisfaction with the system, 2) enjoyment of the interactive experience, 3) naturalness and fluency of responses, 4) accuracy of motion execution, 5) ability to understand human-expressed intentions, and 6) whether the system could bring convenience to their daily lives. Ratings are given on a scale of 1–10, where 10 represents the highest score. A score of 6 represents a baseline level of satisfaction.

Table \ref{tab: DHH_evaluation} presents the individual ratings from all 10 evaluators. Most evaluators give scores above 6 in the majority of categories, indicating general satisfaction with the system’s performance. The convenience criterion received the highest average score (mean: 6.60), suggesting that evaluators think this research is meaningful and has the potential to positively impact their daily lives. If we exclude Evaluators 8 and 9 (likely outliers due to unfamiliar regional signs), the average score across all criteria increases significantly (e.g., Satisfaction average rises from 6.10 to 7.13). Naturalness (from 5.80 to 6.75) and Accuracy (from 5.60 to 6.50) receive moderately high scores, despite the absence of facial expressions or lip movements.

{\bf Limitation.}  Based on feedback from members of the DHH community, we have summarized several current limitations of our framework: 1) {\it Limitations in Sign Language Translation and Generation.} Current state-of-the-art open-source models for sign language translation and generation still exhibit errors. Although we currently use public sign language datasets, our research has found that sign language can vary across different regions within the same country. 2) {\it Limitations of hardware support}. Our robot cannot support facial expressions or lip movements, which may affect the understanding of certain sign language content. 

Despite these limitations, a significant portion of the DHH community strongly affirms both the technical quality of our implementation and the social value of our research. 
\vspace{-0.05in}
\section{Conclusion}\label{sec: conclusion}
\vspace{-0.05in}
We introduce SignBot, a human-robot sign language interaction framework that incorporates an embodied cerebellum + cerebral cooperation mechanism. This framework has been validated across various sign language motions, demonstrating exceptional accuracy, generalization, naturalness, and adaptability across diverse sign language scenarios. In particular, the cerebellum + cerebral cooperation mechanism in SignBot achieves reliable performance in daily communication through the sign language translator, response, and generator. We think SignBot is a foundational solution for sign language applications, such as daily sign language robots serving the DHH community. In the future, we will focus on enabling the robot to exhibit facial expressions, enhancing the interaction experience, and collecting field data in different provinces to accommodate regional variations in sign language.

\section*{APPENDIX}

\section*{A. Experiment Preparation}
\label{sec: Experiment Preparation}
\begin{figure}[htbp]
    \centering
    \includegraphics[width=0.98\linewidth]{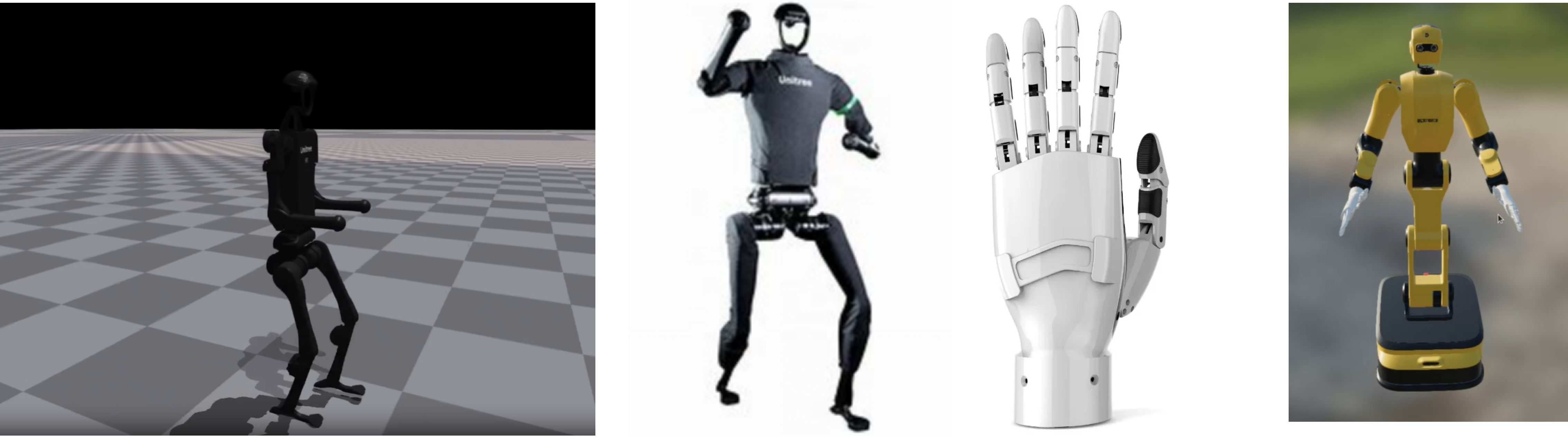}
    \caption{ Experiment Preparation: The above shows the environment for loading the H1/W1 robot in IsaacGym, with the H1/W1 robot in the lower left and the Linker hand dexterous hand in the lower right.}
    \label{fig:product}
\end{figure}

This section mainly introduces the simulation platform we used, the humanoid robot, the dexterous hand, and the Chinese sign language dataset. Figure \ref{fig:product} shows the platform and robotic equipment used in our experiments.

\subsection*{A.1. Experiment Platform}
\vspace{0.1in}\noindent{\bf IsaacGym} is a high-performance robotics simulation and reinforcement learning training platform based on the PhysX physics engine and GPU parallel computing technology. It supports real-time simulation of tens of thousands of robot instances running simultaneously on a single GPU. The platform is optimized for research in robot control, motion planning, and reinforcement learning, providing a native Python interface and an extensible API that allows developers to efficiently train complex policies (such as bipedal robot gaits and robotic arm grasping). Its core advantage lies in its massive parallelization capability, significantly accelerating training efficiency by offloading physics computations to the GPU, and it is widely used in research and development in fields such as industrial automation and human-robot interaction.

\subsection*{A.2. Experiment Equipment}
\vspace{0.1in}\noindent{\bf Unitree H1}: is a full-size general-purpose humanoid robot, weighing approximately 47 kg and standing around 180 cm tall, matching the physical dimensions of an average adult. Its body boasts 19 degrees of freedom (DoF), with bio-inspired joint designs in the legs and high-performance motors. The system is capable of integrating reinforcement learning algorithms to achieve real-time environmental mapping and disturbance-resistant balance control, while also supporting future integration with LLMs to enhance interactive capabilities.

\vspace{0.1in}\noindent{\bf Linker Hand} dexterous hand product has a maximum of 42 degrees of freedom, successfully simulating the fine movements of the human hand. With its 360-degree rotation capability, we have integrated the Linker Hand into humanoid robots to enable them to perform a wider range of tasks in future research work.

\vspace{0.1in}\noindent{\bf W1 Robot} 
The W1 stands 170 cm tall and is equipped with 34 advanced power units, providing strong and stable power support for the robot's operation, ensuring precise and accurate coordination across all functions. Each power unit has a communication bandwidth of up to 100 Mbps. In terms of motion control, the W1 achieves an impressive control frequency of 1000 times per second. It is equipped with a 7-degree-of-freedom humanoid robotic arm, with a maximum single-arm load capacity of 10 kg, sufficient to handle various heavy load tasks. Additionally, the repeat positioning accuracy can be controlled within a precise range of ±0.5 mm. The W1 is also equipped with dual cameras and binocular visual recognition algorithms, which can be used for perception tasks.

\begin{figure}[htbp]
    \centering
    \includegraphics[width=0.98\linewidth]{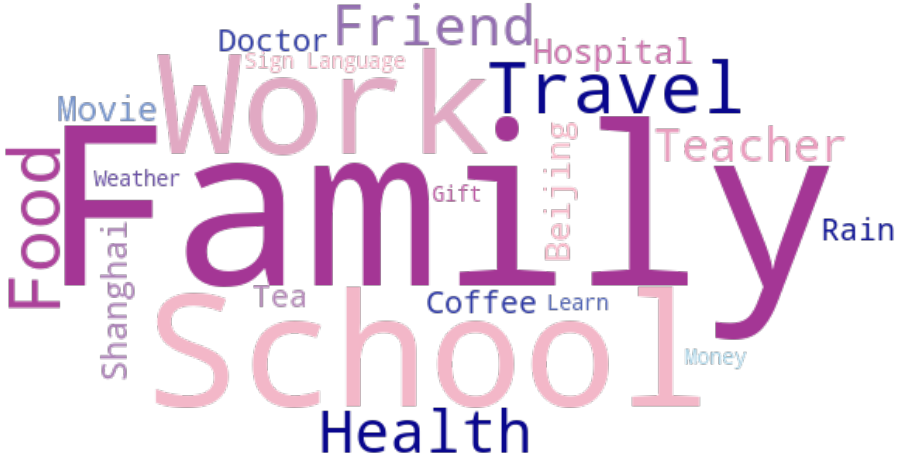}
    \caption{ CSL-Daily Dataset WordCloud Figure.}
    \label{fig:wordcloud}
\end{figure}

\vspace{0.1in}\noindent{\bf CSL-Daily Dataset} is a large-scale continuous Chinese sign language dataset aimed at promoting research and applications in sign language recognition, translation, and generation technologies. This dataset contains over 20,000 sign language videos recorded by 10 sign language users, covering daily life scenarios including family life, school life, and healthcare. Each video is captured using high-definition cameras from various angles with synchronized collection, and each video is meticulously annotated with text. We create a word cloud (see Figure \ref{fig:wordcloud}) to help readers better understand this dataset.

\vspace{0.05in}\noindent{\bf Parametric Human Model} The SMPLX model \cite{Pavlakos2019smplx} based on the SMPL extends parametric human representations with expressive capabilities for body, hands, and face. SMPLX parameterizes the human form through body shape parameters $\beta \in \mathbb{R}^{10}$, joint pose parameters $\theta \in \mathbb{R}^{55\times3}$, and global translation $p \in \mathbb{R}^3$. The SMPLX function $\mathcal{S}(\beta,\theta_p,p): \beta,\theta_p,p \to \mathbb{R}^{10475\times3}$ maps these parameters to the 3D coordinates of a high-fidelity mesh with 10,475 vertices.

\section*{B. Training Details}
\label{sec: Training Details}
This section mainly introduces some of the environment parameters and algorithm parameters we used. During training, we also made special adjustments to certain joints. For example, we do not track the ankle joint to avoid restricting the H1's flexibility in adjusting its position. Additionally, we slightly bent the knee joint to ensure greater stability.

\subsection*{B.1. Task, Regularization and Penalty Rewards}
Table \ref{tab:other_reward} showcases the Task, Regularization, and Penalty Rewards in RL training. 

{\it Task rewards} $r_{t}$ assess how effectively the agents achieve the goals of the current task, such as approaching the target joint Dof pos. These rewards are closely correlated with the optimality of the current control policy, making them an appropriate {\it objective to maximize} in this context. {\it  Penalty rewards} $r_{p}$ aim to prevent undesirable events. For example, these penalties will take effect when the humanoid falls or the robot violates dynamic constraints, including torque limits, joint position limits, and so on. {\it Regularization rewards} $r_{r}$ are used to align the behavior of the humanoid with human preferences regarding motion styles, safe deployment, and other kinematic or dynamic concerns. 

\begin{table}[htbp]
\centering
\resizebox{0.49\textwidth}{!}{
\begin{tabular}{c|c|c}
\hline
Term & Expression & Weight \\
\hline
\multicolumn{3}{c}{Task Reward} \\
\hline
DoF position & $exp{(-0.7 | \hat{\mathbf{q}}_t - \mathbf{q}_t |)}$ & $6.0$ \\
Keypoint position & $exp{(-| \dot{\hat{\mathbf{q}}}_t - \dot{\mathbf{q}}_t |)}$ & $5.0$ \\
Body linear velocity & $exp{(-4.0| \hat{\mathbf{v}}_t - \mathbf{v}_t |)}$ & $6.0$ \\
Body Roll & $exp{(-| \hat{R}_t - R_t |)}$ & $1.0$ \\
Body Pitch & $exp{(-| \hat{P}_t - P_t |)}$ & $1.0$ \\
Body Yaw & $exp{(-| \hat{Y}_t - Y_t |)}$ & $1.0$ \\
\hline
\multicolumn{3}{c}{Penalty Reward} \\
\hline
DoF pos limit & $1(\mathbf{q}_t \notin [\mathbf{q}_{\min}, \mathbf{q}_{\max}])$ & $-1e^{-2}$ \\
Alive & $1$ & $1$ \\
\hline
\multicolumn{3}{c}{Regularization Reward} \\
\hline
Time in Air & $\sum_{i} t^{air}_i * \mathrm{1}^{new \, contact}$ & 10.0 \\
Drag & $\sum_{i} ||\mathbf{v}_{foot}|| * \mathrm{1}^{new \, contact}$ & -0.1 \\
Contact Force & $1\{ |F^{i}_{t}| \geq F_{th} \} * (|F^{i}_{t}| - F_{th})$ & -3e-3 \\
Stumble & $1\{ \exists i, |F^{i}_{xy}| > 4|F^{i}_{t}| \}$ & -2.0 \\
DoF Acceleration & $|\dot{\mathbf{q}}|^2$ & -3e-7 \\
Action Rate & $|\mathbf{a}_{r} - \mathbf{a}_{r}|$ & -1e-1 \\
Energy & $|\dot{\mathbf{q}}|^2$ & -1e-3 \\
DoF Limit Violation & $1 \{ \mathbf{q}_{i} > \mathbf{q}_{max} || \mathbf{q}_{i} < \mathbf{q}_{min} \}$ & -10.0 \\
DoF Deviation & $||\mathbf{q}_{default}^{low}-\mathbf{q}^{low}||^2$ & -1e-1 \\
Vertical Linear Velocity & $\mathbf{v}_{2}$ & -1.0 \\
Horizontal Angular Velocity & $|\mathbf{\omega}_{xy}|^2$ & -0.4 \\
Projected Gravity & $\|\mathbf{g}_{xy}\|^2$ & -2.0\\
\hline
\end{tabular}
}
\caption{Task, Penalty and regularization reward terms.}
\label{tab:other_reward}
\vspace{-0.2in}
\end{table}

\subsection*{B.2. PPO Parameters}
Table \ref{tab:hyperparams} shows the PPO parameters. 

In our PPO implementation, the actor network is composed of a convolutional neural network (CNN) followed by multilayer perceptrons (MLPs), where the CNN is primarily responsible for extracting features from the historical observation inputs. The critic network, in contrast, is constructed purely from MLPs, leveraging the extracted state features for value estimation.
 
\begin{table}[ht]
\centering
\caption{PPO Hyperparameters}
\label{tab:hyperparams}
\begin{tabular}{l|c}
\toprule
Hyperparameter & Value \\
\midrule
Discount Factor & 0.99 \\
GAE Parameter & 0.95 \\
Timesteps per Rollout & 21 \\
Epochs per Rollout & 5 \\
Minibatches per Epoch & 4 \\
Entropy Bonus & 0.01 \\
Value Loss Coefficient & 1.0 \\
Clip Range & 0.2 \\
Reward Normalization & yes \\
Learning Rate & $1\text{e}{-3}$ \\
Optimizer & Adam \\
\bottomrule
\end{tabular}
\end{table}

\subsection*{B.3. Domain Randomization}

Domain randomization is implemented across all training environments. This technique introduces variability in various environmental parameters, including friction and gravity. Detailed specifications of the domain randomization can be found in Table~\ref{tab:sim_params}.

\begin{table}[h]
\centering
\begin{tabular}{c|c}
\hline
Term & Value \\
\hline
\multicolumn{2}{c}{Dynamics Randomization} \\
\hline
Friction & $\mathcal{U}(0.6, 2)$ \\
Base CoM offset & $\mathcal{U}(-0.07, 0.07)$m \\
Base Mass offset & $\mathcal{U}(-1, 5)$kg\\
Motor Strength & $\mathcal{U}(0.8, 1.2) \times$  default\\
Gravity & $\mathcal{U}(-0.1, 0.1)$\\
Link Mass offset& $\mathcal{U}(0.7, 1.3)$kg\\
PD Gains & $\mathcal{U}(0.75, 1.25)\times$  default\\
\hline
\multicolumn{2}{c}{External Perturbation} \\
\hline
Push robot & interval = 8s, $v_{xy} = 0.3$m/s \\
\hline
\end{tabular}
\caption{Domain randomization parameters: we include the friction, gravity, motor strength, base CoM/mass offset of torso link, link mass, PD Gains, and an external perturbation.}
\label{tab:sim_params}

\end{table}

\subsection*{B.4. Running Setting}
We run experiments with three different seeds (123, 321, 1) and present the mean ± std results for each algorithm. The training process for the motion control policy utilizes 12GB of GPU memory across 1 RTX 4090 and typically runs around 48 hours.
\vspace{-0.03in}

 To ensure a fair comparison, we maintain the same settings for all comparison baselines. 
\subsection*{B.5. Torque Setting}
The PD gains, characterized by stiffness and damping values, used in the IssacGym simulator are detailed in Table \ref{tab:torque_params}.

\begin{table}[ht]
\centering
\caption{Torque Parameters}
\label{tab:torque_params}
\resizebox{0.49\textwidth}{!}{
\begin{tabular}{l|ccc}
\toprule
Joint Names & Stiffness [N$\cdot$m/rad] & Damping [N$\cdot$m$\cdot$s/rad] & Torque Limit [Nm] \\
\midrule
hip yaw    & 200    & 5    & 170    \\
hip roll    & 200    & 5    & 170    \\
hip pitch    & 200    & 5    & 170    \\
knee    & 300    & 6    & 255    \\
ankle    & 40    & 2    & 34    \\
torso    & 200    & 5    & 170    \\
shoulder    & 30    & 2    & 34    \\
elbow    & 30    & 2    & 18    \\
hand    & 30    & 2    & 18    \\
\bottomrule
\end{tabular}
}
\end{table}

\subsection*{C. Case Study}
\label{sec:case_study}

\begin{figure}[htbp]
    \centering
    \includegraphics[width=1\linewidth]{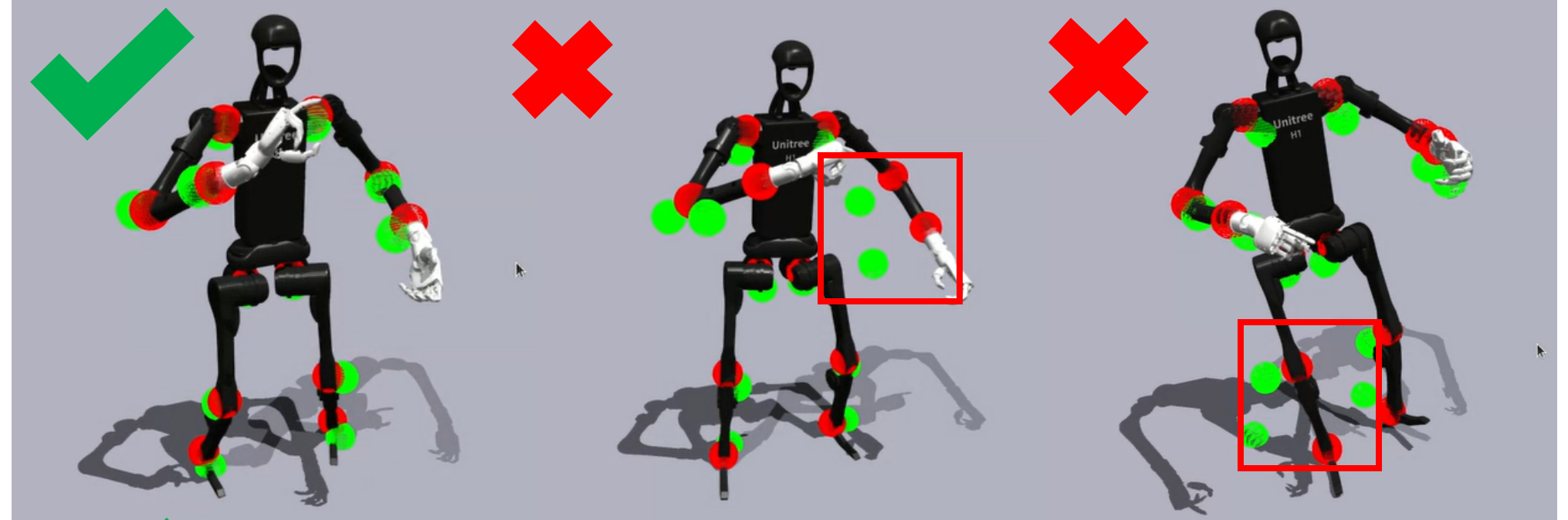}
    \caption{ From left to right: SignBot, Whole-body tracking, and SignBot (w/o Lower-Body Tracking).}
    \label{fig:case_study}
\end{figure}

We conduct a case study (see Figure \ref{fig:case_study}), and it can be observed that while whole-body tracking provides relatively stable tracking of the lower body, the upper body exists some errors due to its inability to keep up with the motion frequency. 
Additionally, if SignBot does not track the posture of the lower body, it becomes very difficult to maintain stability. In addition, because our H1 robot has a pair of larger dexterous hands and a pair of modified wrists, the weight of the upper body has increased. Employing upper body tracking/whole body tracking methods makes it difficult for the robot to maintain its center of gravity while standing still. Therefore, we decouple the policy to allow RL to primarily learn to track the lower body, reducing the difficulty of training.

\bibliographystyle{IEEEtran}
\bibliography{IEEEabrv}


\end{document}